\pdfoutput=1

\documentclass[11pt]{article}

\usepackage[review]{ACL2023}

\usepackage{times}
\usepackage{latexsym}

\usepackage[T1]{fontenc}

\usepackage[utf8]{inputenc}

\usepackage{microtype}

\usepackage{inconsolata}
\usepackage{graphicx}
\usepackage{eso-pic}
\usepackage{xcolor}
\usepackage{amsmath}
\usepackage{amssymb}
\usepackage{cite}
\usepackage{amsfonts}
\usepackage{algorithmic}
\usepackage{subcaption}
\usepackage{multirow}
\usepackage{booktabs}
\usepackage{bm}
\usepackage{color}
\usepackage{algorithm}
\usepackage{xspace}
\usepackage{soul}
\usepackage{enumitem}
\usepackage{textcomp}
\usepackage{scalerel}
\usepackage{makecell}
\usepackage{pifont}
\usepackage{bbding}
\usepackage{colortbl}
\usepackage{array}
\usepackage{url}
\usepackage{hhline}

\title{AMBER: An LLM-free Multi-dimensional Benchmark for MLLMs Hallucination Evaluation}

\author{
  Junyang Wang$^{1}$\thanks{\;\;Equal contribution}\;\,\thanks{\, Work done during internship at Alibaba Group.}\;\;, Yuhang Wang$^{1*}$, Guohai Xu$^2$, Jing Zhang$^1$, Yukai Gu$^1$, \\\bf Haitao Jia$^1$, Jiaqi Wang$^1$, Haiyang Xu$^2$, Ming Yan$^2$, Ji Zhang$^2$, Jitao Sang$^{1,3}$\thanks{\, Corresponding author}\\
  $^1$Beijing Jiaotong University\quad
  $^2$Alibaba Group\quad
  $^3$Peng Cheng Lab\\
  \texttt{\tt\small 
  \{junyangwang, yhangwang, zhangj\_, yukai.gu, jiaht1101, jiaqiw, jtsang\}@bjtu.edu.cn}\\
  \texttt{\tt\small 
  \{guohai.xgh, shuofeng.xhy, ym119608, zj122146\}@alibaba-inc.com
  }
  }
\nolinenumbers  
\begin{document}

\definecolor{darkgreen}{rgb}{0.0, 0.5, 0.0}
\definecolor{lightgray}{rgb}{0.9, 0.9, 0.9}
\definecolor{Mycolor1}{HTML}{BAD8F2}
\definecolor{Mycolor2}{HTML}{E0F0FA}

\maketitle

\begin{abstract}
    Despite making significant progress in multi-modal tasks, current Multi-modal Large Language Models (MLLMs) encounter the significant challenge of hallucinations, which may lead to harmful consequences. 
    Therefore, evaluating MLLMs' hallucinations is becoming increasingly important in model improvement and practical application deployment.
    Previous works are limited in high evaluation costs (e.g., relying on humans or advanced LLMs) and insufficient evaluation dimensions (e.g., types of tasks and hallucinations). 
    In this paper, we propose an LLM-free multi-dimensional benchmark AMBER, which can be used to evaluate both generative task and discriminative task including existence, attribute and relation hallucination.
    Based on AMBER, we design a low-cost and efficient evaluation pipeline.
    Additionally, we conduct a comprehensive evaluation and detailed analysis of mainstream MLLMs including GPT-4V(ision), and also give guideline suggestions for mitigating hallucinations.
    The data and code of AMBER are available at \url{https://github.com/junyangwang0410/AMBER}.
\end{abstract}

\section{Introduction}

\begin{table*}[t]
    \centering
    \renewcommand{\arraystretch}{1.1}
    \setlength{\tabcolsep}{10pt}
    \scalebox{0.9}{
    \begin{tabular}{l|cc|c}
    \toprule 
     \textbf{Evaluation Method} & \textbf{Task Type} & \textbf{Hallucination Type} & \textbf{LLM-free}\\
     \midrule
     POPE~\citep{li2023evaluating} & Discriminative  & Exis.& \color{darkgreen}\Checkmark \\ 
     M-HalDetect~\citep{gunjal2023detecting} & Generative  & N/A & \color{red}\ding{55} \\ 
     HaELM~\citep{wang2023evaluation}  & Generative & N/A & \color{red}\ding{55} \\ 
     Halle-Switch~\citep{zhai2023halle} & Generative  & N/A & \color{red}\ding{55} \\
     \rowcolor{lightgray}
     AMBER~(ours) &  Generative\&Discriminative &  Exis. \& Attr. \& Rel. & \color{darkgreen}\Checkmark \\ 
    \bottomrule 
  \end{tabular}}
    \caption{Comparison with existing hallucination evaluation methods.
    We utilize the following abbreviations:
    Exis.- Existence; Attr.-Attribute; Rel.-Relation.}
    \label{tab:method}
\end{table*} 

GPT-4 exhibited a range of remarkable abilities in vision-language tasks~\citep{OpenAI2023GPT4TR}.
Subsequently, many works developed Multi-modal Large Language models (MLLMs) with similar capabilities to GPT-4 by integrating visual encoders into Large Language Models (LLMs), e.g.,
MiniGPT-4, LLaVA and mPLUG-Owl~\citep{liu2023visual,zhu2023minigpt,ye2023mplug,dai2023instructblip,liu2023improved,chen2023minigpt}. 
However, MLLMs occasionally generate content that seems plausible but is unfaithful to the given image, which is called hallucination~\citep{liu2023aligning,ye2023cognitive,yin2023woodpecker}. 
This may lead to harmful consequences, especially when users without sufficient domain knowledge over-rely on these models.

\begin{figure}[t]
    \centering
    \includegraphics[width=0.45\textwidth]{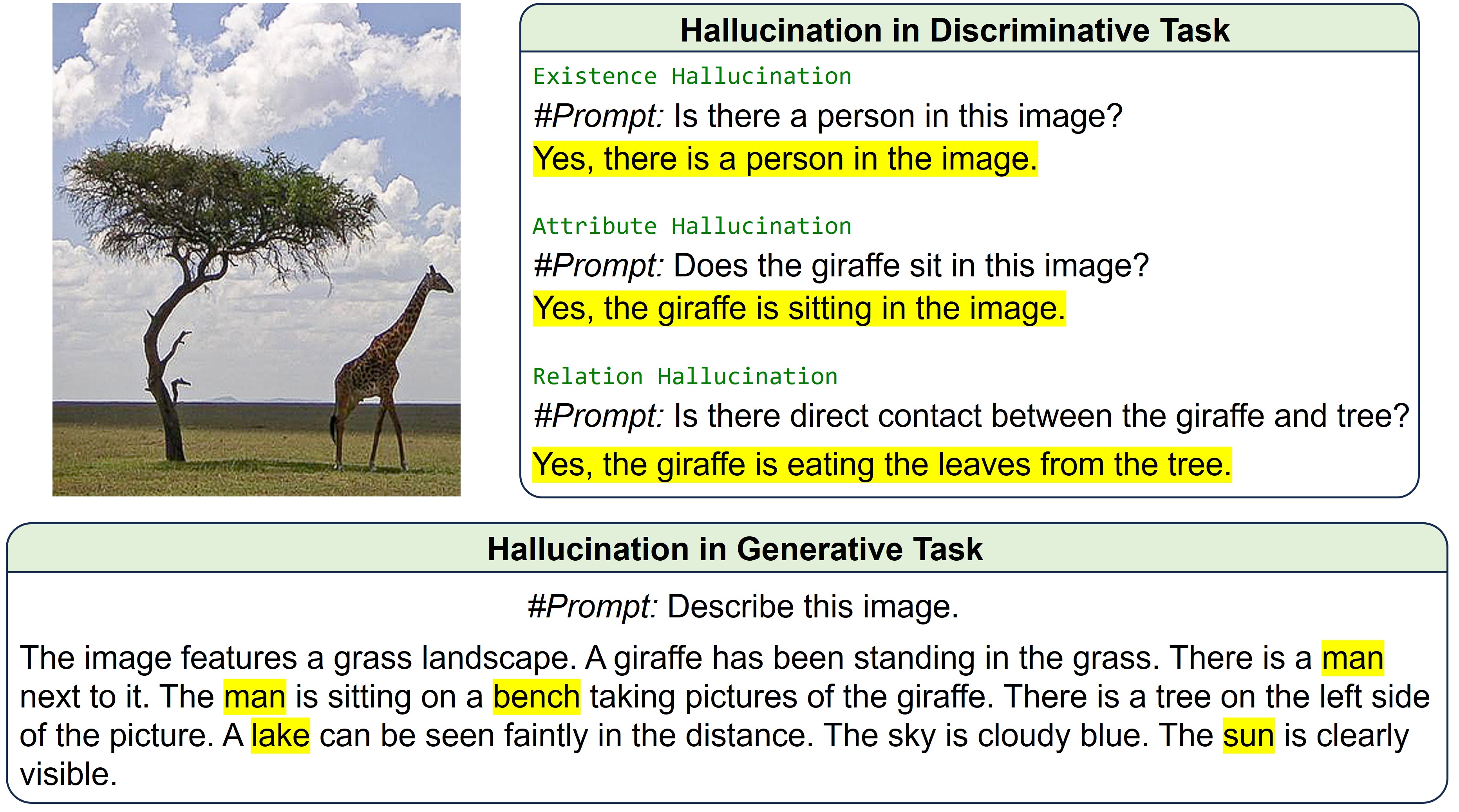}
    \caption{Examples of hallucinations in the generative task and discriminative task of MLLMs. The \hl{highlight} fonts represent the hallucinatory content.}
    \label{fig:example}
\end{figure}

Therefore, evaluating MLLMs' hallucinations is an important step in the actual application deployment and the subsequent enhancement of reliability. 
The hallucinations of MLLMs mainly occur in two major types of task: generative and discriminative,
as shown in the Figure~\ref{fig:example}.
However, there are several limitations of existing evaluation methods for each type of task. 
For the generative task, most evaluation methods suffer from a limitation wherein they rely on additional LLMs. 
This reliance results in a huge cost, posing challenges for scalable applications.
For example, \citet{zhou2023analyzing,zhai2023halle} evaluated MLLMs relying on humans or GPT-4.
\citet{wang2023evaluation,gunjal2023detecting} proposed to train a hallucinatory detection model for evaluation. 
Regarding the discriminative task, \citet{li2023evaluating} proposed an object existence hallucination evaluation method, but lacked evaluation on other types of hallucinations such as 
those related to attributes and relations. 
Note that there is no method incorporating all the above types of tasks and hallucinations. 
We summarize these methods in Table~\ref{tab:method}. 

To address the aforementioned problems, we propose AMBER (\textbf{A}n LLM-free \textbf{M}ulti-dimensional \textbf{Be}nchma\textbf{r}k) for MLLMs hallucination evaluation. First, we collect a batch of high-quality images with clear content and well-defined objects, which have not been used for training MLLMs. 
Then, we provide comprehensive annotations to facilitate the evaluation of both generative and discriminative tasks, 
covering three types of hallucination: existence, attribute and relation, as shown in Figure~\ref{fig:evaluation pipeline}.
Accordingly, we develop an LLM-free evaluation pipeline (see Figure~\ref{fig:evaluation pipeline}).
Based on our proposed AMBER, we conduct multi-dimensional hallucination evaluations on nine mainstream MLLMs, including GPT-4V, and observe persistent hallucination issues.
Furthermore, we investigate the characteristics of hallucinations in MLLMs, 
their contributing factors, and offer effective mitigation strategies based on our findings, such as enriching the structure and distribution of training data, using more powerful visual or language models, controlling the response length, etc.

We summarize the contributions as follows:
\begin{itemize} 
\item We meticulously construct a benchmark AMBER for evaluating hallucinations in both the generative task and discriminative task of MLLMs. 
AMBER provides comprehensive coverage of evaluations for various types of hallucination, including those of existence, attribute and relation.
\item We develop an LLM-free evaluation pipeline based on AMBER, that caters to both types of task and different hallucination types.
\item  Employing AMBER, we evaluate and analyze the mainstream MLLMs including the most advanced GPT-4V. And we give guideline suggestions for mitigating hallucinations.
\end{itemize}

\section{Related Work}
\subsection{MLLMs}

Large Language Models (LLMs), represented by GPT-4, are currently demonstrating powerful language processing capabilities, greatly propelling the development of natural language processing~\citep{OpenAI2023GPT4TR,touvron2023llama,touvron2023llama2,chiang2023vicuna}. In order to apply the formidable capabilities of LLMs to multi-modal tasks, researchers have integrated visual modules with LLMs, giving rise to Multi-modal Large Language Models (MLLMs). MLLMs achieve multitask learning in the multi-modal domain through instruction learning and showcase robust capabilities in many traditional multi-modal tasks, such as visual question answering, image captioning and prompt-based object detection~\citep{liu2023visual,zhu2023minigpt,ye2023mplug,dai2023instructblip,liu2023improved,chen2023minigpt,ye2023mplugowl2}.
 
\begin{figure*}[t]
    \centering
    \includegraphics[width=0.9\textwidth]{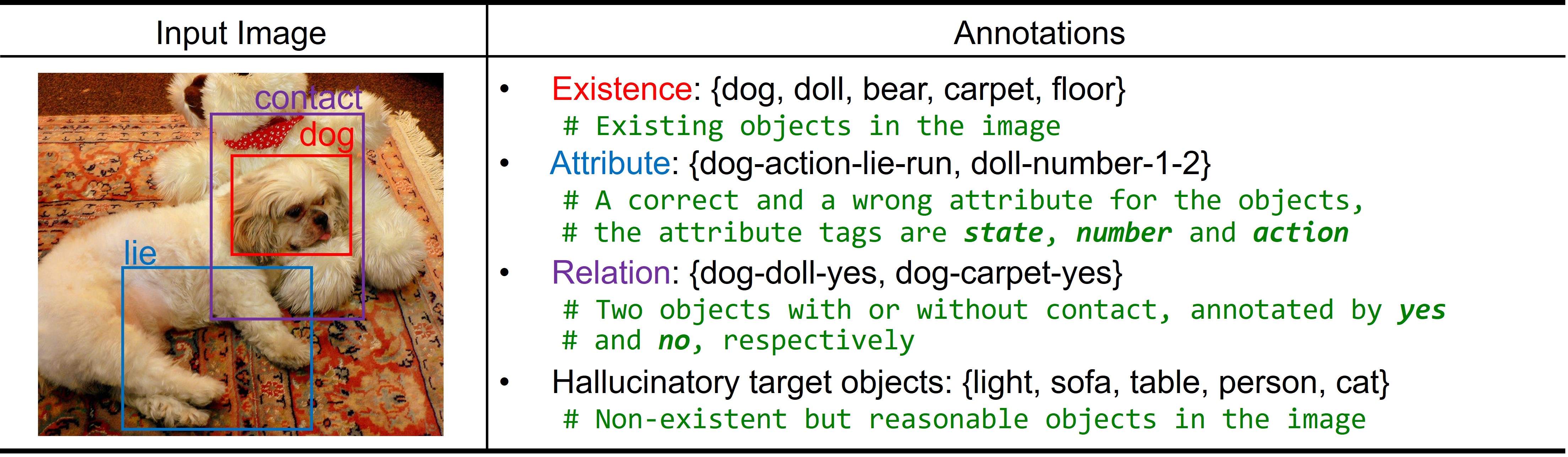}
    \caption{An example of the annotations of AMBER, where the boxes in the images represent the basis for annotating the corresponding content.}
    \label{fig:annotations}
\end{figure*}

\subsection{Hallucination in MLLMs}

While MLLMs have exhibited excellent performance, the MLLMs often generate content in their responses that is unfaithful to the given images, which is called hallucination~\citep{liu2023aligning}. Hallucinations are believed to occur for the following reasons: (1) Insufficiently diverse training data with some errors~\citep{liu2023aligning}; (2) MLLMs losing attention to the image when generating the responses~\citep{wang2023evaluation}; (3) Information loss in visual input after the image is fed into the visual encoder, making it unable to completely align with the textual description~\citep{zhai2023halle}. Hallucinations often occur as descriptions or judgments that are reasonable but not faithful to the images, making them highly deceptive and misleading. Therefore, the presence of hallucinations significantly restricts the application scenarios.

There are currently three main types of methods for evaluating hallucinations in MLLMs. The first is based on human or GPT-4~\citep{zhou2023analyzing,zhai2023halle}. This type of method is relatively reliable but comes with a high cost. This cost becomes particularly prominent in academic research where the effectiveness of experimental setups often requires multiple evaluations. The second is based on hallucinatory detection models~\citep{wang2023evaluation,gunjal2023detecting}. In this type of method, hallucinatory image descriptions are first collected, and then a model is trained to determine whether a given description contains hallucinations. However, the model's performance is highly dependent on hallucinatory data. Additionally, the substantial training costs and inapplicability to discriminative task also exist. The third method is based on object detection~\citep{li2023evaluating}. It involves using detectors to extract all the objects in an image and then asking the MLLMs whether a specific object exists in the image. The MLLMs' response of ``yes'' or ``no'' is used to determine the presence of hallucinations. This method is only applicable to discriminative task and only evaluate the existence hallucination.

To address the issues of high cost, requirement of LLM and incomplete evaluations of the types of task and hallucination in the existing methods, we introduce AMBER, an LLM-free multi-dimensional benchmark.

\section{The AMBER Benchmark}

In this section, we introduce the process of constructing AMBER and the pipeline for evaluating MLLMs' hallucinations.

\subsection{Dataset Construction}

\begin{figure*}[t]
    \centering
    \includegraphics[width=0.9\textwidth]{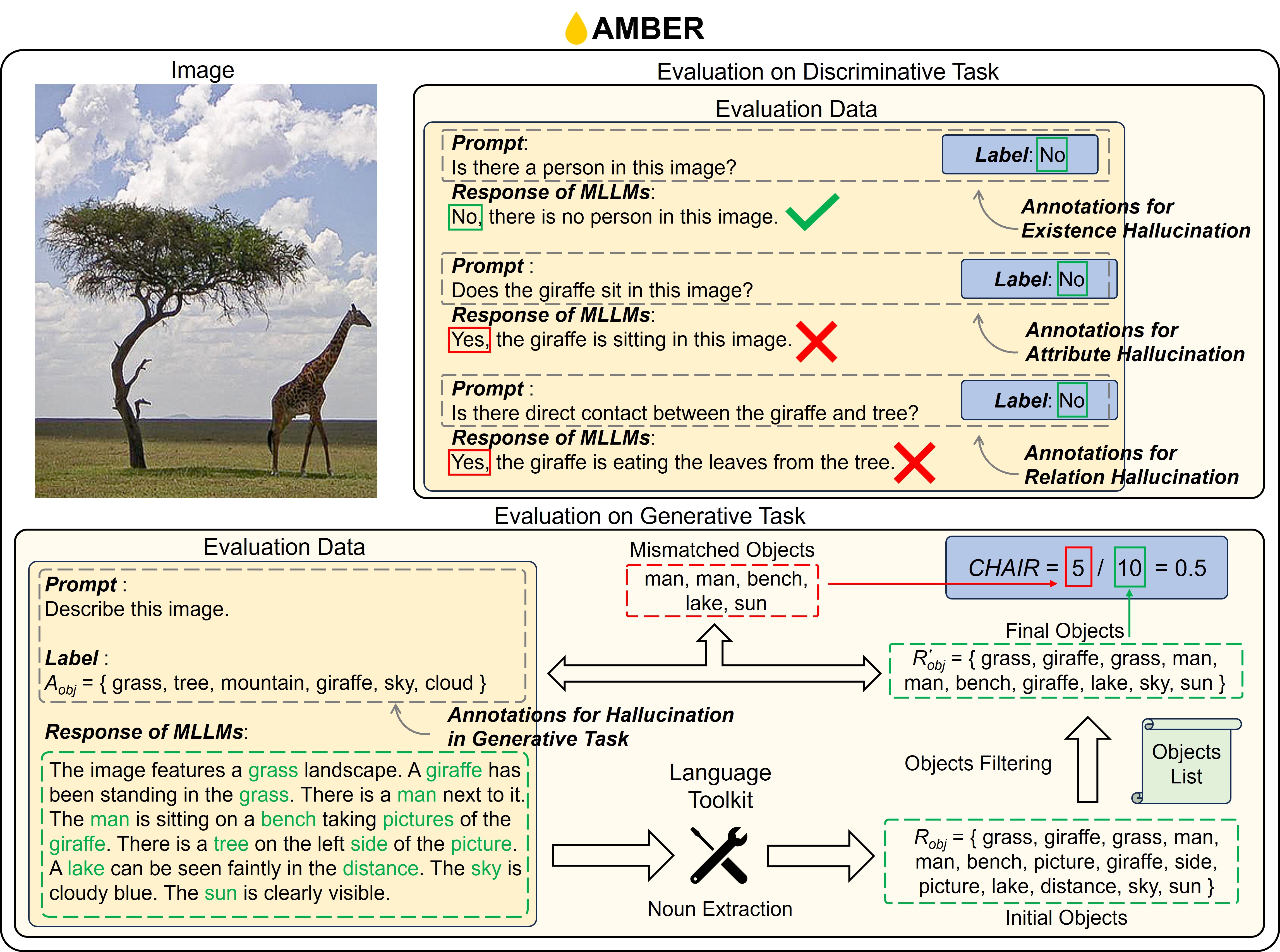}
    \caption{An illustration demonstrating the process of evaluating MLLMs' hallucinations based on AMBER.}
    \label{fig:evaluation pipeline}
\end{figure*}

The construction of the dataset consists of three steps: collecting images, annotating images and designing prompt templates. We will detail each of these steps below.\\
\noindent \textbf{Collecting Images.} Collecting high-quality and diverse images is the foundational step in our work to objectively evaluate the hallucinations of MLLMs.
Existing works indicated that mainstream annotated datasets have been widely used for MLLMs training, resulting in an inflated performance of trained MLLMs on benchmarks composed of such data~\citep{zhou2023don}. 
To avoid this, we source images from test sets of multi-modal open-source datasets and copyright-free image repositories on the Internet.
For image quality, 
we manually select images with clear content and well-defined objects. 
During the annotation phase, images too challenging for accurate annotation are discarded to avoid incomplete annotations.
Additionally, we employed multi-dimensional object filtering to ensure the diversity of the images as far as possible.
After filtering, the total number of images in our dataset is 1,004. 
Detailed data statistics are shown in Appendix~\ref{sec:Details of AMBER}.\\
\noindent \textbf{Annotating Images.} In this step, we meticulously annotate 4 types of content for each image. 
We take an example in Figure~\ref{fig:annotations} to illustrate these annotations. 
\begin{itemize}
\item \textit{Existence} refers to all visible objects in the image, including background objects like the ``carpet'' and ``floor''.
\item \textit{Attribute} refers to the attributes of the existing objects in the image. We annotate them from three perspectives: state (such as color, shape, etc.), number (typically marked when an object appears more than once in the image) and action (actions of humans or animals in the image). 
For example, in ``\textit{dog-action-lie-run}'',  ``\textit{dog}'' is the object,   ``\textit{action}'' indicates the type of attribute,  ''lie`` and ''run`` represent the correct and incorrect annotations of ``\textit{action}'' respectively.
Note that we set up incorrect annotations to construct counterfactual prompts (see \textbf{Designing Prompt Templates}).  
\item  \textit{Relation} refers to whether there is direct contact between two objects in an image.
\item  \textit{Hallucinatory target objects} explicitly do not exist in the image, but they are likely to be imagined by MLLMs based on the image.
\end{itemize}

\par During the annotation phase, multiple groups of annotators label the collected image.
We then coordinate the results to achieve a unified annotation set. 
To reduce potential mistakes from annotators,  we manually inspect the annotation outcomes and correct any erroneous annotations during the annotation validation process.\\
\noindent \textbf{Designing Prompt Templates.} In the last step, we design the prompt templates for evaluation on both generative and discriminative task.
For the generative task, we use the most commonly used generative prompt, ``Describe this image.'', to obtain descriptions of images from MLLMs.\par
For the discriminative task, we design corresponding prompts based on different types of hallucinations. 
We use three prompts ``Is the \{\textit{object}\} \{\textit{state}\} in this image?'', ``Is/Are there \{\textit{number}\} \{\textit{object}\} in this image?'' and ``Does the \{\textit{object}\} \{\textit{action}\} in this image?'' to evaluate state, number and action perspectives of the attribute hallucination, respectively.
For relation hallucination, 
we use the prompt ``Is there direct contact between the \{\textit{object 1}\} and \{\textit{object 2}\} in this image?''. 
For the hallucinatory target objects, we construct the counterfactual prompt ``Is there a \{\textit{object}\} in this image?''. 

\subsection{Evaluation} 
In this subsection, we introduce the evaluation pipeline for AMBER, as shown in Figure~\ref{fig:evaluation pipeline}. We first describe how to obtain and process the responses from MLLMs, then clarify the metrics.
\subsubsection{Prompting MLLMs}
\noindent \textbf{MLLMs Respondents.}
we select nine mainstream MLLMs for evaluation, including LLaVA~\citep{liu2023visual}, MiniGPT-4~\citep{zhu2023minigpt}, mPLUG-Owl~\citep{ye2023mplug}, InstructBlip~\citep{dai2023instructblip}, LLaVA-1.5~\citep{liu2023improved}, Qwen-VL~\citep{bai2023qwen}, CogVLM~\citep{wang2023cogvlm}, mPLUG-Owl2~\citep{ye2023mplugowl2} and GPT-4V~\citep{OpenAI2023GPT4TR}.\par
\noindent \textbf{Response Processing.} 
AMBER provides a set of $Input$=\{$Img$, $Ins$\}, where $Img$ represents the image, $Ins$ denotes the instruction constructed according to
prompt templates and the annotations of the image.
As shown in the Figure~\ref{fig:evaluation pipeline},
we obtain the initial response $R$ by fitting $Input$ into MLLM.\par
Regarding the generative task, 
we first extract nouns from $R$ by language toolkits (e.g., nltk) to obtain the initial objects $R_{obj}$=\{$obj^R_1$, $obj^R_2$, $\ldots$, $obj^R_n$\}.
Then, we construct an objects list $X_{obj}$=\{$obj^X_1$, $obj^X_2$, $\ldots$, $obj^X_n$\} consisting of all annotated objects in AMBER to filter unnecessarily objects in $R_{obj}$ like ``picture'', ``distance'' and ``side'' as shown in Figure~\ref{fig:evaluation pipeline}.
Finally we obtain the final objects $R^{'}_{obj}$ by the follow formula:
\begin{equation}
    R^{'}_{obj} = R_{obj} \cap X_{obj}.
\end{equation}
For the discriminative task, since the queries are in the form of a true/false, we only need to determine if the responses contain ``yes'' or ``no''.

\subsubsection{Metrics.}
\noindent \textbf{Metrics on the Generative Task.} We introduce four metrics for evaluating hallucinations on the generative task.\par
\textit{CHAIR}~\citep{rohrbach2018object} is a commonly used metric for evaluating hallucinations. 
It measures the frequency of hallucinatory objects appearing in the responses. 
AMBER provides an annotated objects list $A_{obj}$=\{$obj^A_1$, $obj^A_2$, $\ldots$, $obj^A_n$\} as seen in Figure~\ref{fig:evaluation pipeline}. 
\textit{CHAIR(R)} is represented by the following formula:
\begin{equation}
    \text{\textit{CHAIR(R)}} = 1 - \frac{len(R^{'}_{obj} \cap A_{obj})}{len(R^{'}_{obj})}.
\end{equation}

\textit{Cover} measures the object coverage of responses. Specifically, it quantifies the proportion of objects mentioned in the response $R^{'}_{obj}$ relative to the objects identified in the $A_{obj}$. 
An ideal response is considered to be one that minimizes hallucinatory content without significantly compromising the coverage of objects in the image.
\textit{Cover(R)} is represented by the following formula:
\begin{equation}
    \text{\textit{Cover(R)}} = \frac{len(R^{'}_{obj} \cap A_{obj})}{len(A_{obj})}.
\end{equation}

\textit{Hal} represents the proportion of responses with hallucinations. 
For a response \textit{R}, if \textit{CHAIR(R)} $\neq$ 0, then \textit{R} is considered to have hallucinations. \textit{Hal(R)} is represented by the following formula:
\begin{equation}
    \text{\textit{Hal(R)}} = \begin{cases}
  1 & \text{if $\text{\textit{CHAIR(R)}} \neq 0$}, \\
  0 & \text{otherwise} .
\end{cases}
\end{equation}

\textit{Cog.}
To assess whether the hallucinations in MLLMs are similar to those in human cognition, we utilize a set of hallucinatory target objects, denoted as $H_{obj}$=\{$obj^H_1$, $obj^H_2$, $\ldots$, $obj^H_n$\}.
These objects are employed to develop a metric named \textit{Cog} for evaluating the likelihood of MLLMs generating the objects from $H_{obj}$. 
\textit{Cog(R)} is represented by the following formula:
\begin{equation}
    \text{\textit{Cog(R)}} = \frac{len(R^{'}_{obj} \cap H_{obj})}{len(R^{'}_{obj})}.
\end{equation}

Note that we calculate the average values of the four mentioned metrics over all the queries in AMBER, represented by \textit{CHAIR}, \textit{Cover}, \textit{Hal}, and \textit{Cog}, respectively.

\begin{table*}[t]
    \centering
    \renewcommand{\arraystretch}{1.1}
    \setlength{\tabcolsep}{9pt}
    \scalebox{0.88}{
    \begin{tabular}{l | c c c c | c c c c | c}
    \hline
    \toprule
    \multicolumn{1}{l}{\multirow{2}{*}{\textbf{Model}}}&\multicolumn{4}{c}{\textsc{Generative Task}}&\multicolumn{4}{c}{\textsc{Discriminative Task}}&\multicolumn{1}{c}{\multirow{2}{*}{\textbf{\makecell{AMBER \\ Score}}}}\\
    \cmidrule(lr){2-5}
    \cmidrule(lr){6-9}
    \multicolumn{1}{c}{}&\multicolumn{1}{c}{CHAIR$_{\downarrow}$}&Cover$_{\uparrow}$&Hal$_{\downarrow}$&\multicolumn{1}{c}{Cog$_{\downarrow}$}&\multicolumn{1}{c}{Acc.}&P.&R.&\multicolumn{1}{c}{F1}&\multicolumn{1}{c}{}\\
    \midrule
    mPLUG-Owl&21.6&50.1&76.1&11.5&40.1&92.8&10.5&18.9&48.7\\
    LLaVA&11.5&51.0&48.8&5.5&42.7&74.1&21.0&32.7&60.6\\
    MiniGPT-4&13.6&\colorbox{Mycolor2}{63.0}&65.3&11.3&63.6&90.5&50.4&64.7&75.6\\
    
    CogVLM&5.6&57.2&\colorbox{Mycolor1}{23.6}&\colorbox{Mycolor1}{1.3}&69.0&88.9&60.9&72.3&83.4\\
    LLaVA-1.5&7.8&51.0&36.4&4.2&72.0&\colorbox{Mycolor2}{93.2}&62.4&74.7&83.5\\
    mPLUG-Owl2&10.6&52.0&39.9&4.5&75.6&\colorbox{Mycolor1}{95.0}&66.9&78.5&84.0\\
    InstructBLIP&8.8&52.2&38.2&4.4&76.5&84.5&79.0&81.7&86.5\\
    Qwen-VL&\colorbox{Mycolor2}{5.5}&49.4&\colorbox{Mycolor1}{23.6}&\colorbox{Mycolor2}{1.9}&\colorbox{Mycolor2}{81.2}&90.8&\colorbox{Mycolor2}{79.7}&\colorbox{Mycolor2}{84.9}&\colorbox{Mycolor2}{89.7}\\
    GPT-4V&\colorbox{Mycolor1}{4.6}&\colorbox{Mycolor1}{67.1}&\colorbox{Mycolor2}{30.7}&2.6&\colorbox{Mycolor1}{83.4}&84.9&\colorbox{Mycolor1}{90.1}&\colorbox{Mycolor1}{87.4}&\colorbox{Mycolor1}{91.4}\\
    \bottomrule
    \hline
    \end{tabular}
    }
    \caption{The overall evaluation results of AMBER on generative task and discriminative task.}
    \label{tab:overall evaluation results}
\end{table*} 
 
\noindent \textbf{Metrics on Discriminative Task.}
Since MLLMs are constrained to answer with a binary ``yes'' or ``no'' on discriminative task, 
we use standard classification metrics: \textit{Accuracy}, \textit{Precision}, \textit{Recall} and \textit{F1}. 
As AMBER is designed for hallucination evaluation, both \textit{Precision} and \textit{Recall} are calculated under the hallucinatory questions (where the ground truth is ``no''). 
To prevent MLLMs from being biased towards rejection and obtaining high scores, \textit{Accuracy} is still calculated for all questions.

\noindent \textbf{\textit{AMBER Score.}}
To comprehensively evaluate the performance of various MLLMs under both types of task, we introduce the \textit{AMBER Score} to integrate the \textit{CHAIR} on generative task and the \textit{F1} on discriminative task. The \textit{AMBER Score} is represented by the following formula:
\begin{equation}
    \text{\textit{AMBER Score}} = \frac{1}{2}\times(1 - \textit{CHAIR} + \textit{F1}).
\end{equation}

\section{Results}

\subsection{AMBER Overall Results}
As shown in Table~\ref{tab:overall evaluation results}, 
the performance of nine models in terms of hallucination is presented in ascending order based on \textit{AMBER Score}.
Notably, GPT-4V stands out as the top performer, exhibiting impressive results in both generative and discriminative tasks.
However, GPT-4V still presents some instances of hallucinations, which we will discuss in detail in section~\ref{sec:gpt-4v}.
In second place is Qwen-VL, which demonstrates a lower incidence of hallucinations in both task types and excels in three out of four metrics.
Additionally, we observe that both mPLUG-Owl2 and LLaVA-1.5 show significant improvements over their first-generation counterparts, mPLUG-Owl and LLaVA, respectively. 
We speculate that this might be due to the upgraded versions utilizing stronger foundational LLMs and the enhanced quality of training data.
Furthermore, MLLMs tend to produce more hallucinations in discriminative task compared to the generative task, and we will introduce them separately below.\par
\noindent \textbf{Results on Generative Task.}
We observe that various MLLMs generally exceed a \textit{Hal} value of 30\% and 
even the most advanced MLLMs still have around 5\% hallucinatory objects when facing open-ended generative instructions.
This indicates that reducing hallucinations in the generative task remains a significant challenge.
CogVLM and Qwen-VL demonstrate significant advantages with high values of \textit{CHAIR}, \textit{Hal} and \textit{Cog}.
MiniGPT-4 excels in the \textit{Cover} but with more hallucinations. 
The high \textit{Cog} indicates that MiniGPT-4 is more prone to associations and also to make mistakes.\par
\noindent \textbf{Results on Discriminative Task. \label{sec:Results on Discriminative Task}}
We find that the \textit{recall} values for all MLLMs are significantly lower than the \textit{precision} values.
This indicates that MLLMs tend to give affirmative responses, suggesting that they are easily misled by the hallucinatory content of the questions, as shown in \citep{li2023evaluating}.
We find that Qwen-VL and InstructBLIP exhibit a significant advantage in discriminative task.
Actually, they also perform well in a similar VQA-v2~\citep{balanced_vqa_v2} benchmark. 
We believe an effective way to mitigate hallucinations is to enhance MLLMs' capabilities on discriminative benchmarks such as VQA-v2.

\begin{table*}[t]
    \centering
    \renewcommand{\arraystretch}{1.1}
    \setlength{\tabcolsep}{6.5pt}
    \scalebox{0.9}{
    \begin{tabular}{l | c c c | c c c c | c c c c}
    \toprule
    \multicolumn{1}{l}{\multirow{2}{*}{\textbf{Model}}}&\multicolumn{3}{c}{\textsc{Existence}}&\multicolumn{4}{c}{\textsc{Attribute}}&\multicolumn{4}{c}{\textsc{Relation}}\\
    \cmidrule(lr){2-4}
    \cmidrule(lr){5-8}
    \cmidrule(lr){9-12}
    \multicolumn{1}{c}{}&\multicolumn{1}{c}{P.}&R.&\multicolumn{1}{c}{F1}&\multicolumn{1}{c}{Acc.}&P.&R.&\multicolumn{1}{c}{F1}&\multicolumn{1}{c}{Acc.}&P.&R.&F1\\
    \midrule
    mPLUG-Owl&99.7&9.4&17.2&55.7&87.5&13.2&22.9&59.6&\colorbox{Mycolor2}{81.5}&3.2&6.2\\
    LLaVA&\colorbox{Mycolor2}{99.9}&4.4&8.4&62.9&78.9&35.1&48.6&63.8&55.7&60.8&58.1\\
    MiniGPT-4&\colorbox{Mycolor2}{99.9}&66.7&80.0&61.7&82.7&29.7&43.7&63.4&56.7&49.3&52.7\\
    CogVLM&\colorbox{Mycolor1}{100}&73.3&84.5&66.8&79.7&44.9&57.4&66.7&59.8&59.9&59.8\\
    LLaVA-1.5&\colorbox{Mycolor1}{100}&71.5&83.3&72.0&\colorbox{Mycolor2}{88.0}&51.0&64.6&\colorbox{Mycolor2}{73.9}&72.0&60.2&65.6\\
    mPLUG-Owl2&\colorbox{Mycolor1}{100}&80.4&89.1&76.6&\colorbox{Mycolor1}{88.1}&61.5&72.4&58.6&\colorbox{Mycolor1}{81.8}&40.6&54.3\\
    InstructBLIP&\colorbox{Mycolor1}{100}&80.2&89.0&76.1&75.9&76.7&76.3&66.8&56.7&\colorbox{Mycolor2}{83.6}&\colorbox{Mycolor2}{67.6}\\
    Qwen-VL&\colorbox{Mycolor1}{100}&\colorbox{Mycolor2}{83.5}&\colorbox{Mycolor2}{91.0}&\colorbox{Mycolor1}{81.9}&83.7&\colorbox{Mycolor2}{79.2}&\colorbox{Mycolor2}{81.4}&71.3&69.1&55.4&61.6\\
    GPT-4V&\colorbox{Mycolor1}{100}&\colorbox{Mycolor1}{89.6}&\colorbox{Mycolor1}{94.5}&\colorbox{Mycolor2}{80.4}&75.2&\colorbox{Mycolor1}{90.6}&\colorbox{Mycolor1}{82.2}&\colorbox{Mycolor1}{84.0}&77.1&\colorbox{Mycolor1}{90.3}&\colorbox{Mycolor1}{83.2}\\
    \bottomrule
    \end{tabular}
    }
    \caption{
The detailed evaluation results on discriminative task, where \textsc{Existence}, \textsc{Attribute} and \textsc{Relation} represent existence hallucination, attribute hallucination and relation hallucination, respectively. Since the the questions used to evaluate the existence hallucination are all negative responses, recall and accuracy are equivalent.}
    \label{tab:experiment12}
\end{table*}

\subsection{Multi-dimensional Results}
Benefiting from the fine-grained annotations of AMBER, we conduct a multi-dimensional analysis on existence, attribute and relation hallucination on discriminative tasks. We will introduce them one by one below.\par
\noindent \textbf{Existence Hallucination.} The results in Table~\ref{tab:experiment12} indicate that current mainstream MLLMs are not heavily troubled by the existence hallucination. 
However, we can see that all the \textit{precision} values are close to 1, but the \textit{recall} values don't match accordingly. 
The common issue that MLLMs are more inclined to provide affirmative responses inspires the idea of incorporating more negative samples during the training phase.\par
\noindent \textbf{Attribute Hallucination.} In contrast, attribute hallucination poses more significant challenges. Even the best-performing MLLM achieves only an \textit{F1} value of around 0.8. 
We conduct a detailed analysis of attribute hallucination from perspectives of state, number and action in Appendix~\ref{Detailed Results of Attribute Hallucination}.\par
\noindent \textbf{Relation Hallucination.} MLLMs are most susceptible to induction and hallucination in terms of relation. 
The \textit{accuracy} values of most MLLMs are only around 0.7. 
This may stem from the fact that current multi-modal instruction fine-tuning datasets offer limited detail about the relationships between objects, instead focusing more on individual objects.
Therefore, we suggest data creators enrich the object descriptions from attribute and relation perspectives.

\subsection{GPT-4V(ision) \label{sec:gpt-4v}}
Existing works demonstrate that GPT-4V also produces a few hallucinations in certain scenarios~\citep{liu2023hallusionbench,wu2023early,cui2023holistic}. 
Our evaluations using AMBER, confirm these findings, and we illustrate some specific examples of hallucinations in Appendix~\ref{sec:gpt-4v}.
However, GPT-4V stands out as the most robust among current MLLMs.
As detailed in Table~\ref{tab:overall evaluation results},  GPT-4V not only presents the least hallucinations but also achieves the highest image object coverage in the generative task, according to the \textit{CHAIR} and \textit{Cover}.
This highlights the advanced capabilities of GPT-4V, as it surpasses the existing MLLMs in the trade-off between hallucinations and coverage.
Regarding discriminative task, GPT-4V continues to make a strong impression, as indicated in Table~\ref{tab:overall evaluation results}.
Table~\ref{tab:experiment12} further shows that GPT-4V achieved the best performance across all three types of hallucination.

\begin{figure}[t]
    \centering
    \includegraphics[width=0.4\textwidth]{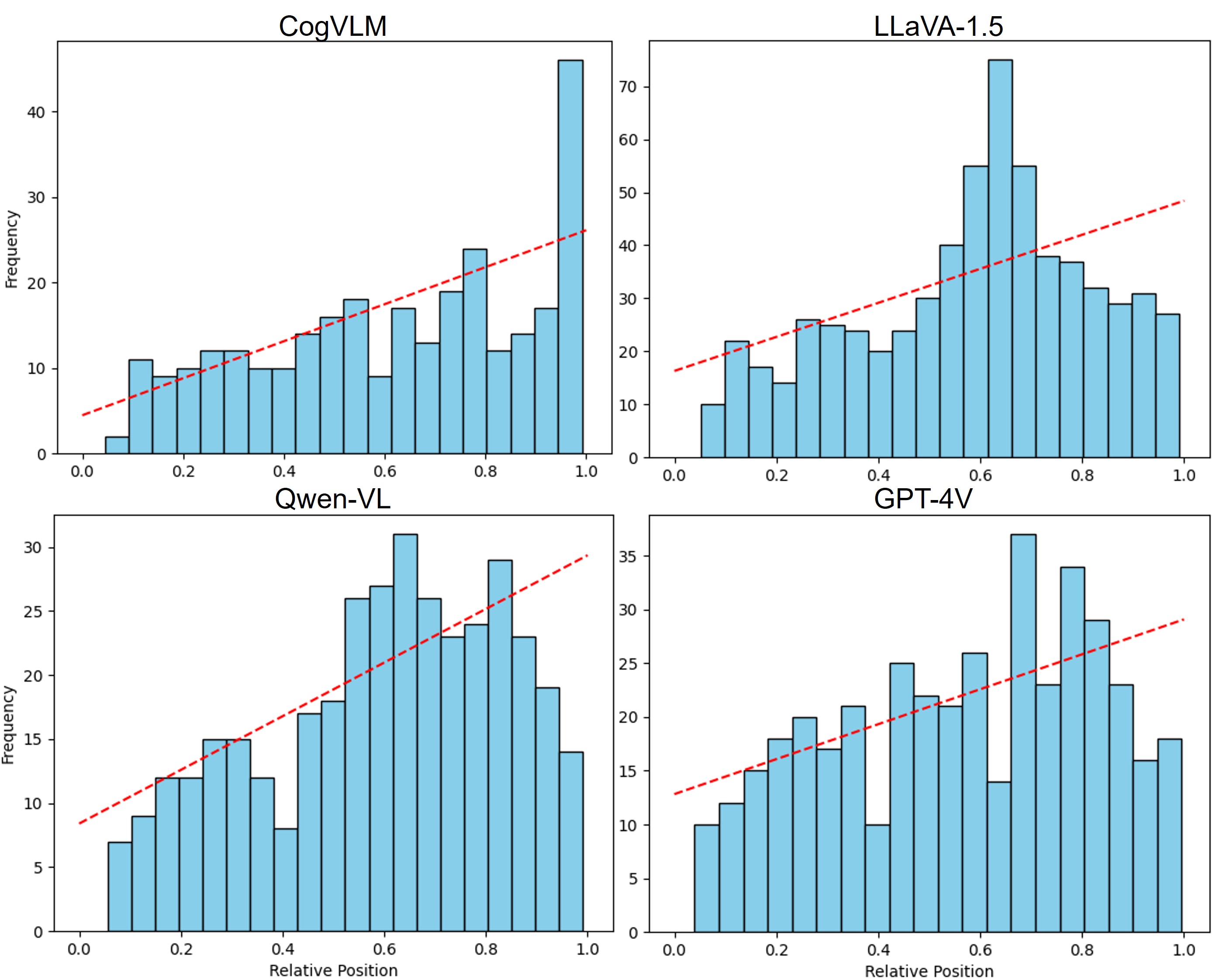}
    \caption{The frequency of hallucinatory objects appearing at different relative positions in MLLMs' responses.}
    \label{fig:position}
\end{figure}

\subsection{Analysis of Hallucinations in MLLMs}
\noindent \textbf{Position of Hallucination Appearance.}
We focus on the generative task and select four models with relatively low levels of hallucination to investigate where hallucinatory objects tend to appear.
The experimental results are depicted in Figure \ref{fig:position}. 
It is evident that hallucinatory objects frequently occur in the middle and latter parts of the responses,
which may be attributable to snowballing or error accumulation.\par
\begin{figure}[t]
    \centering
    \includegraphics[width=0.41\textwidth]{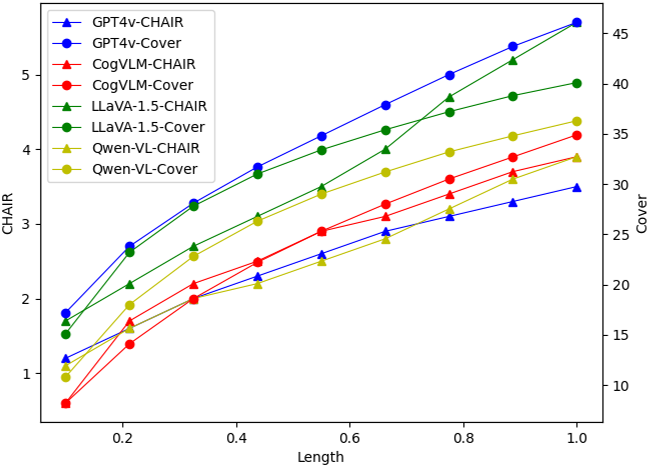}
    \caption{Trends of the metrics \textit{Hal} and \textit{Cover} for different MLLMs with respect to the relative response length.}
    \label{fig:len}
\end{figure}

\noindent \textbf{Hallucination vs. Object Coverage.} 
To analyze the relationship between the hallucination and object coverage, we truncate the responses at different lengths and calculate \textit{Cover} and \textit{Hal} to evaluate object coverage and hallucination, respectively. 
As shown in Figure~\ref{fig:len}, 
there is a noticeable trade-off trend that, as the response length increases, both object coverage and hallucinations tend to increase correspondingly.
While more conservative responses might reduce hallucinations, it is important to ensure that this does not compromise the quality of the response.\par
\noindent \textbf{Tendency to Respond \textit{Yes} or \textit{No}.}
A notable challenge for MLLMs is their propensity to produce affirmative responses. 
This tendency occurs when responding to prompts in discriminative task, irrespective of whether these responses accurately reflect the content of the images.
We compile the response tendencies of MLLMs as shown in Table~\ref{tab:yesno}. 
The results show that the \textit{Yes Ratio} values of almost all MLLMs are higher than \textit{No Ratio}. 
Only InstructBLIP does not show a clear tendency, while Qwen-VL and GPT-4V tend to provide negative responses. 
The response tendencies of MLLMs significantly influence their propensity to generate erroneous content. Therefore, it is crucial to incorporate negative responses into training data to reduce such hallucinations.
\begin{table}[t]
  \centering
  \renewcommand{\arraystretch}{1.1}
    \setlength{\tabcolsep}{6pt}
    \scalebox{0.9}{
  \begin{tabular}{l | c c c c} 
  \hline
    \toprule 
    \multicolumn{1}{l}{\multirow{2}{*}{\textbf{Model}}}&\multicolumn{2}{c}{Yes Ratio}&\multicolumn{2}{c}{No Ratio}\\
    \cmidrule(lr){2-3}
    \cmidrule(lr){4-5}
    \multicolumn{1}{c}{}&\multicolumn{1}{c}{All.}&w/ Hal.&All.&w/o Hal.\\
    \midrule
    CogVLM&71.8&27.5&28.2&5.7\\
    LLaVA-1.5&71.0&24.5&29.0&3.5\\
    mPLUG-Owl2&69.6&22.3&28.2&4.3\\
    InstructBLIP&49.5&11.7&50.5&12.2\\
    Qwen-VL&41.8&12.2&58.2&20.5\\
    GPT-4V&39.7&4.7&60.3&15.0\\
    \bottomrule
    \hline
  \end{tabular}}
  \caption{The ratio of MLLMs' responses with ``Yes'' or ``No'' on discriminative tasks, where w/ Hal and w/o Hal represent the ground truth for queries with ``Yes'' and ``No'', respectively.}
  \label{tab:yesno}
  \vspace{-0.3cm}
\end{table}

\subsection{Ablation Study} \label{ablation_section}
\noindent \textbf{Visual Resolution.} 
Previous studies have demonstrated that enhancing image resolution can lead to improved performance. 
We modify the image resolution of AMBER to investigate its impact on hallucinations. 
The results are shown in Table~\ref{tab:resolution}. 
Following the increase in resolution,
there is a decrease in the level of hallucinations.
It is worth noting that GPT-4V is minimally affected by resolution. 
This also illustrates the robustness of GPT-4V in terms of hallucination for image resolution.\par
\begin{table}[!ht]
  \centering
  \renewcommand{\arraystretch}{1.1}
    \setlength{\tabcolsep}{6pt}
    \scalebox{0.9}{
  \begin{tabular}{l | c c c c} 
  \hline
    \toprule 
    \multicolumn{1}{l}{\multirow{2}{*}{\textbf{Model}}}&\multicolumn{4}{c}{Relative Resolution}\\
    \cmidrule(lr){2-5}
    \multicolumn{1}{c}{}&\multicolumn{1}{c}{25\%}&50\%&75\%&100\%\\
    \midrule
    CogVLM&83.0&83.1&83.3&83.4\\
    LLaVA-1.5&82.9&83.1&83.5&83.5\\
    Qwen-VL&89.1&89.3&89.6&89.7\\
    GPT-4V&91.2&91.4&91.3&91.4\\
    \bottomrule
    \hline
  \end{tabular}}
  \caption{\textit{AMBER Score} variation with image resolution changes.}
  \label{tab:resolution}
\end{table}
\noindent \textbf{The Scales of LLMs.} 
We investigate the impact of hallucinations by varying the scales of LLMs, 
and the results are shown in Table~\ref{tab:llm_size}.
As the number of parameters in the LLM increases, 
MiniGPT-4 shows a rising tendency for hallucinations in the generative task and a reduced incidence in discriminative tasks. 
In contrast, LLaVA-1.5 shows an opposing trend. 
The impact of LLM's scales on these two MLLMs does not demonstrate a consistent pattern.
This variance suggests that hallucination formation in MLLMs might be more influenced by their vision or connectivity modules.\par
\begin{table}[!ht]
  \centering
  \renewcommand{\arraystretch}{1.1}
    \setlength{\tabcolsep}{6pt}
    \scalebox{0.85}{
  \begin{tabular}{l|cccc} 
  \hline
    \toprule
    \multicolumn{1}{l}{\multirow{2}{*}{\textbf{Model}}}&\multicolumn{2}{c}{Generative}&\multicolumn{2}{c}{Descriminative}\\
    \cmidrule(lr){2-3}
    \cmidrule(lr){4-5}
    \multicolumn{1}{c}{}&\multicolumn{1}{c}{CHAIR$_{\downarrow}$}&Cover$_{\uparrow}$&Acc&F1\\
    \midrule
    MiniGPT-4 7B&\colorbox{Mycolor1}{13.6}&63.0&63.6&64.7\\
    MiniGPT-4 13B&14.3&\colorbox{Mycolor1}{63.6}&\colorbox{Mycolor1}{64.9}&\colorbox{Mycolor1}{66.7}\\
    \midrule
    LLaVA-1.5 7B&7.8&\colorbox{Mycolor1}{51.0}&72.0&\colorbox{Mycolor1}{74.4}\\
    LLaVA-1.5 13B&\colorbox{Mycolor1}{7.0}&50.8&\colorbox{Mycolor1}{72.2}&73.1\\
    \bottomrule
    \hline
  \end{tabular}}
  \caption{Results of MLLMs under different scales of LLMs.}
  \label{tab:llm_size}
  \vspace{-0.3cm}
\end{table}
\noindent \textbf{Training Data.}
Academic datasets and multi-modal instruction data have been widely used in the training phase of MLLMs. 
To assess their impact on producing hallucinations, we evaluate MiniGPT-v2 in both stage-2 and stage-3.
In stage-2, the model was trained on academic datasets, 
During stage-3, it received additional training with the incorporation of multi-modal instructional data
The results presented in Table~\ref{tab:stage} indicate that stage-3 exhibits fewer hallucinations on the generative task 
but shows an increase in discriminative task. 
This phenomenon can be attributed to the structured output format of academic datasets, which is more conducive to enhancing the model's discriminative capabilities.
While the addition of multi-modal instruction data strengthens the model's generative capacity, 
it also introduces challenges to discriminative abilities in an open-ended generation.
\begin{table}[!ht]
  \centering
  \renewcommand{\arraystretch}{1.1}
    \setlength{\tabcolsep}{7pt}
    \scalebox{0.9}{
  \begin{tabular}{l|cccc} 
  \hline
    \toprule
    \multicolumn{1}{l}{\multirow{2}{*}{\textbf{\makecell{Training \\ Data}}}}&\multicolumn{2}{c}{Generative}&\multicolumn{2}{c}{Descriminative}\\
    \cmidrule(lr){2-3}
    \cmidrule(lr){4-5}
    \multicolumn{1}{c}{}&\multicolumn{1}{c}{CHAIR$_{\downarrow}$}&Cover$_{\uparrow}$&Acc&F1\\
    \midrule
    Stage-2&35.0&22.1&\colorbox{Mycolor1}{37.5}&\colorbox{Mycolor1}{16.6}\\
    Stage-3&\colorbox{Mycolor1}{32.8}&\colorbox{Mycolor1}{31.6}&36.2&10.8\\
    \bottomrule
    \hline
  \end{tabular}}
  \caption{Results of MiniGPT-v2 in different stages.}
  \label{tab:stage}
  \vspace{-0.3cm}
\end{table}

\section{Conclusion}

MLLMs are rapidly advancing, but the hallucinations still require attention. Existing methods for hallucination evaluation suffer from high costs and insufficient evaluation dimensions. In this work, we introduce AMBER, an LLM-free multi-dimensional benchmark for the hallucination evaluation of MLLMs. AMBER does not rely on additional LLMs and evaluates across various tasks and hallucination types. We conducted a detailed analysis of mainstream MLLMs by AMBER, exploring factors related to hallucinations and providing suggestions for mitigation. In future work, we will delve deeper into mitigating hallucinations and advance the reliability of MLLMs.

\section{Limitations}

The AMBER proposed in this work offers significant advantages in terms of evaluation efficiency, cost and quality. However, AMBER still has some limitations. Firstly, while AMBER provides evaluation algorithms for attribute and relation hallucination, these evaluations are still limited to discriminative task. This is because we cannot guarantee that MLLMs can provide effective descriptions of attributes and relations in the generative task. Secondly, there is a possibility of mistakes in extracting the object using language toolkits. For example, ``orange'' can be used as an adjective to describe color, but we find that nltk might identify it as a noun. Similar situations also include compound nouns, for example, the ``water'' of ``water bottle'' may be mistakenly considered as a hallucination object. Lastly, the hallucinatory target objects we annotated are still limited, which means that less common objects in images may be overlooked. As MLLMs' recognition capabilities improve, the number of hallucinatory target objects should also increase accordingly.

\bibliography{custom}
\bibliographystyle{acl_natbib}
\clearpage
\appendix
\section{Appendix}
\label{sec:appendix}

\subsection{Details of AMBER \label{sec:Details of AMBER}}

\subsubsection{Data Source}

Our images are sourced from the MS-COCO 2014~\citep{lin2014microsoft} test set and UnSplash\footnote{www.unsplash.com}. It is important to note that the MS-COCO 2014 test set does not include annotations and is therefore not used as training data. UnSplash adopts the Creative Commons Zero (CC0) license, allowing free use anywhere without requiring attribution to the source or author.

\subsubsection{Data Dtatistics}

We present the data statistics for AMBER in Table~\ref{tab:statistics}. The category column represents the number of different categories present in the data. It is worth mentioning that the resultant annotations cover 337 objects, which is more than 4 times the number of objects in the existing benchmarks (e.g., 80 specific objects in coco).

\begin{table}[!ht]
  \centering
  \begin{tabular}{llc} 
  \hline
    \toprule 
    \multicolumn{2}{l}{\textbf{Image}} & 1004 \\ 
    \midrule
    \multirow{2}{*}{\textbf{Categories}} 
     & Object  & 337 \\
     & Attribute  & 350 \\
    \midrule
    \multirow{4}{*}{\textbf{Prompt}} 
    & Generation  & 1004 \\
      & Existence  & 4924 \\
      & Attribute  & 7628 \\
      & Relation  & 1664 \\
    \bottomrule
    \hline
  \end{tabular}
  \caption{Data Statistics.}
  \label{tab:statistics}
\end{table}

\subsubsection{Object Distribution}

\begin{figure}[!ht]
    \centering
    \includegraphics[width=0.48\textwidth]{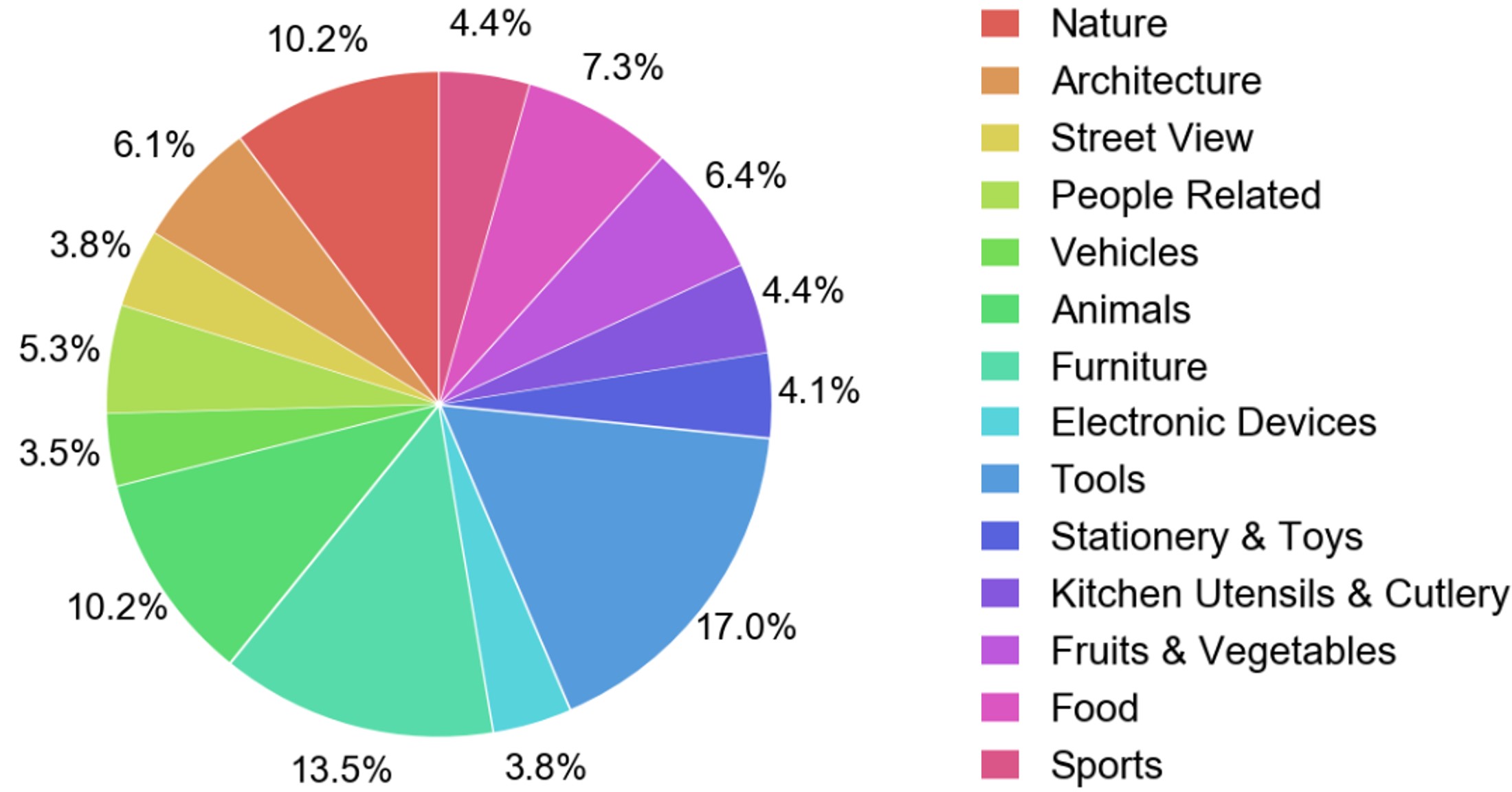}
    \caption{The distribution of objects in AMBER.}
    \label{fig:distribution}
\end{figure}

In Figure~\ref{fig:distribution}, we present the distribution of detectable objects in AMBER. These objects cover 14 major categories, and the distribution of each category is relatively balanced, without a significant long-tail phenomenon. Compared to existing benchmarks, AMBER achieves the first coverage of objects in Nature, Architecture, and Street View. Additionally, in other categories, AMBER achieves maximum extension, such as in the Fruit category, where the annotations support the detection of over a dozen common fruits compared to existing methods that can only detect three types of fruits.

\subsection{MLLMs Respondents}

Table~\ref{tab:model} shows the detailed configurations of the MLLMs we evaluated. These MLLMs include both early models like LLaVA and recent ones like Qwen-VL. For fairness, we used the official hyperparameters provided for generation for each MLLM. Responses from MLLMs were not truncated based on length. It is worth noting that we did not evaluate MiniGPT-v2~\citep{chen2023minigpt}. This is because MiniGPT-v2 uses tags to control the response format, which we consider unfair as additional information for other models. We plan to evaluate more MLLMs in future work to create a comprehensive list.
\begin{table*}[!ht]
    \centering
    \begin{tabular}{l | c c | c c}
    \hline
    \toprule    
    \textbf{Model}&\textbf{Vision Encoder} & \textbf{\#Params.} & \textbf{Language Model}& \textbf{\#Params.}\\
    \midrule
    LLaVA~\citep{liu2023visual}& ViT-L/14 & 0.4B & Llama-2-Chat & 7B \\
    MiniGPT-4~\citep{zhu2023minigpt}& ViT-G/14 & 2.0B &Vicuna & 7B\&13B \\ 
    mPLUG-Owl~\citep{ye2023mplug}& ViT-L/14 & 0.4B & Llama & 7B \\
    InstructBLIP~\citep{dai2023instructblip}& ViT-G/14 & 2.0B &Vicuna & 7B\\
    LLaVA-1.5~\citep{liu2023improved}& ViT-L/14-336px & 0.4B & Vicuna-v1.5  & 7B\&13B\\
    Qwen-VL~\citep{bai2023qwen}&ViT-G/14&2.0B&Qwen&7B\\
    CogVLM~\citep{wang2023cogvlm}& EVA2-CLIP-E & 4.7B & Vicuna-v1.5 & 7B \\
    mPLUG-Owl2~\citep{ye2023mplugowl2}& ViT-L/14 & 0.4B & Llama-2-Chat & 7B\\
    GPT-4V~\citep{OpenAI2023GPT4TR} & Unknown & Unknown & GPT-4 & Unknown\\
    \bottomrule
    \hline
    \end{tabular}
    \caption{The architecture and parameters of MLLMs evaluated by AMBER.}
    \label{tab:model}
\end{table*}

\begin{table*}[!ht]
    \centering
    \renewcommand{\arraystretch}{1.2}
    \setlength{\tabcolsep}{5pt}
    \scalebox{0.9}{
    \begin{tabular}{l | c c c c | c c c c | c c c c}
    \hline
    \toprule
    \multicolumn{1}{l}{\multirow{2}{*}{\textbf{Model}}}&\multicolumn{4}{c}{\textsc{State}}&\multicolumn{4}{c}{\textsc{Number}}&\multicolumn{4}{c}{\textsc{Action}}\\
    \cmidrule(lr){2-5}
    \cmidrule(lr){6-9}
    \cmidrule(lr){10-13}
    \multicolumn{1}{c}{}&\multicolumn{1}{c}{Acc.}&P.&R.&\multicolumn{1}{c}{F1}&\multicolumn{1}{c}{Acc.}&P.&R.&\multicolumn{1}{c}{F1}&\multicolumn{1}{c}{Acc.}&P.&R.&F1\\
    \midrule
    mPLUG-Owl&57.2&86.4&17.1&28.5&50.5&\colorbox{Mycolor1}{92.3}&1.2&2.4&60.0&92.5&21.7&35.2\\
    LLaVA&64.9&76.2&43.5&55.4&57.1&\colorbox{Mycolor2}{91.6}&15.7&26.8&65.3&87.1&35.9&50.8\\
    MiniGPT-4&67.8&84.7&43.3&57.3&49.9&48.7&3.8&7.0&56.3&85.7&15.2&25.8\\
    CogVLM&64.7&76.7&42.1&54.4&64.6&79.8&39.0&52.4&85.1&91.4&77.5&83.9\\
    LLaVA-1.5&68.5&\colorbox{Mycolor2}{86.6}&43.8&58.2&75.4&87.7&59.1&70.6&84.1&\colorbox{Mycolor2}{93.8}&73.0&82.1\\
    mPLUG-Owl2&75.2&\colorbox{Mycolor1}{87.4}&59.4&70.5&76.6&89.1&60.6&72.1&85.6&\colorbox{Mycolor1}{94.0}&76.3&84.1\\
    InstructBLIP&76.0&73.5&\colorbox{Mycolor2}{81.4}&77.2&72.4&78.9&61.2&68.9&\colorbox{Mycolor2}{86.7}&85.4&\colorbox{Mycolor2}{88.6}&\colorbox{Mycolor2}{87.0}\\
    Qwen-VL&\colorbox{Mycolor1}{79.3}&81.7&75.4&\colorbox{Mycolor2}{78.4}&\colorbox{Mycolor2}{85.7}&86.1&\colorbox{Mycolor2}{85.1}&\colorbox{Mycolor2}{85.6}&\colorbox{Mycolor1}{87.8}&88.8&86.4&\colorbox{Mycolor1}{87.6}\\
    GPT-4V&\colorbox{Mycolor2}{76.5}&73.1&\colorbox{Mycolor1}{88.9}&\colorbox{Mycolor1}{79.1}&\colorbox{Mycolor1}{88.4}&85.6&\colorbox{Mycolor1}{92.5}&\colorbox{Mycolor1}{88.9}&83.0&76.4&\colorbox{Mycolor1}{95.5}&84.9\\
    \bottomrule
    \hline
    \end{tabular}
    }
    \caption{The detailed evaluation results of AMBER on attribute hallucination.}
    \label{tab:experiment13}
\end{table*}

\subsection{Results}

\subsubsection{Detailed Results of Attribute Hallucination \label{Detailed Results of Attribute Hallucination}}

We dissected the attribute hallucination of AMBER into three perspectives—state, number, and action, obtaining refined evaluations for attribute hallucination, as presented in Table~\ref{tab:experiment13}.

In terms of state and number, MLLMs generally perform poorly. discriminative task related to state and number investigate MLLMs' fine-grained examination of both single and multiple objects. We recommend that data creators focus on the state and number details of objects.

In terms of action, we find that MLLMs generally perform well. It is noteworthy that, in this perspective, the performance of currently available open-source MLLMs is better than that of GPT-4V. 

We believe this may be related to the bias in the training data. Existing multi-modal instruction datasets often have richer descriptions of object action, while information about the states and number of objects is relatively scarce. We suggest data creators enhance coverage of more attribute dimensions for the objects.

\subsubsection{Analysis of Hallucination}
\noindent \textbf{Tendency to Respond \textit{Yes} or \textit{No}} \par
For the tendency of discriminative task, we have added the results for mPLUG-Owl, MiniGPT-4, and LLaVA to the Table~\ref{tab:yesno_all}. These MLLMs perform relatively poorly on AMBER. From the results, it can be seen that they exhibit a stronger inclination towards positive responses, with over 30\% being affirmative replies to hallucination-related questions, while non-hallucination questions receive less than 5\% negative responses. The excessively strong tendency toward positive responses makes MLLMs highly susceptible to the induction of hallucinations in questions, leading to incorrect replies.\\

\noindent \textbf{Preference for Hallucinatory Objects} \par

\begin{figure}[!ht]
    \centering
    \includegraphics[width=0.48\textwidth]{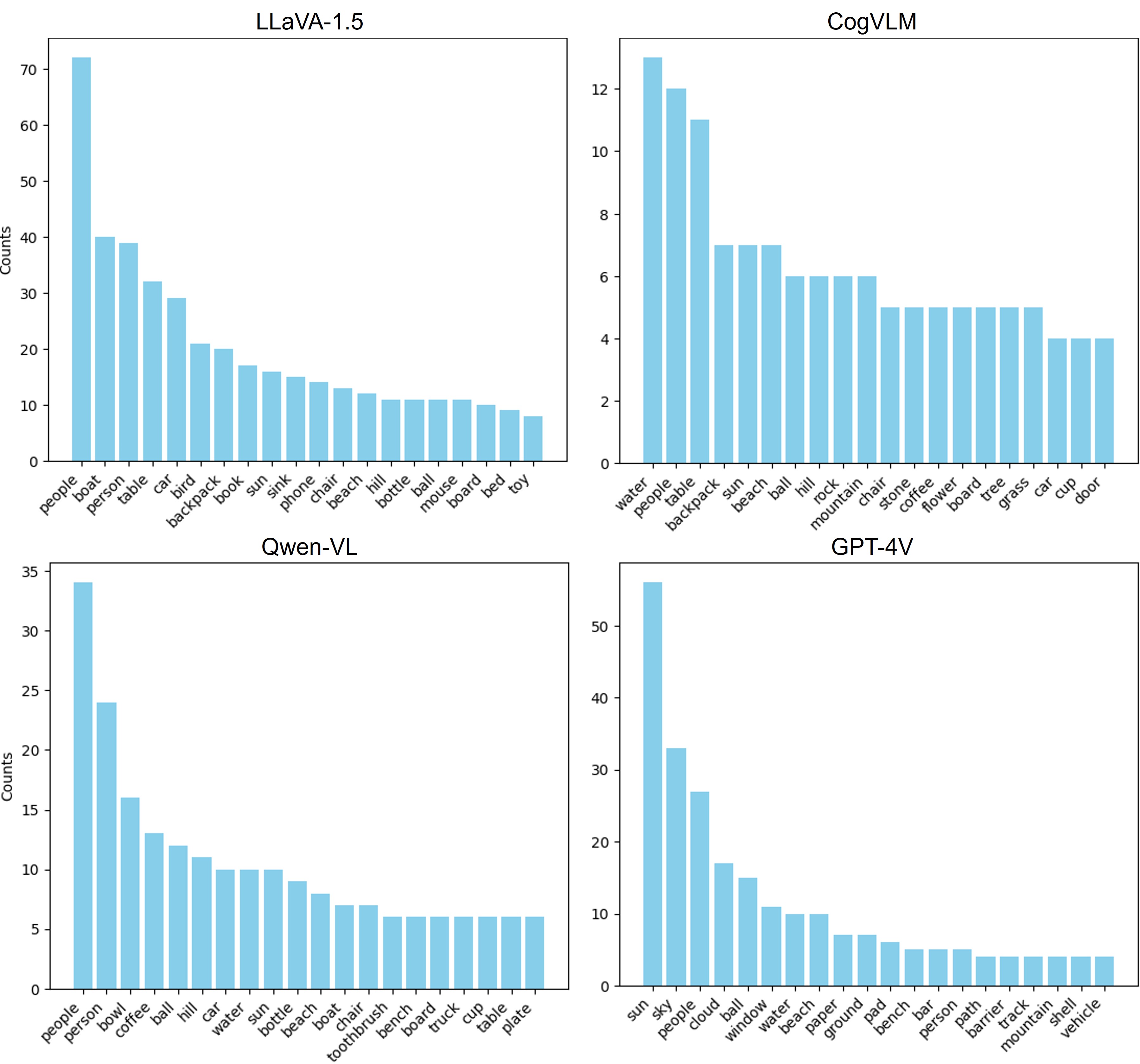}
    \caption{Statistics of the top 20 hallucinatory objects of various MLLMs.}
    \label{fig:prefer}
\end{figure}

Figure~\ref{fig:prefer} shows the preferences of multiple MLLMs in generating hallucinatory objects on generative task. It can be observed that each MLLM has its own preferences, for example, open-source MLLMs tend to generate hallucinations of people, while GPT-4V tends to hallucinate background objects such as the sky, sun, and clouds. We believe that the generation of hallucinations is closely related to the context of the images. Hallucinatory objects are often influenced by certain objects in the images, and errors occur during the embedding of visual modalities into the text modality.

\begin{figure}[!ht]
    \centering
    \includegraphics[width=0.48\textwidth]{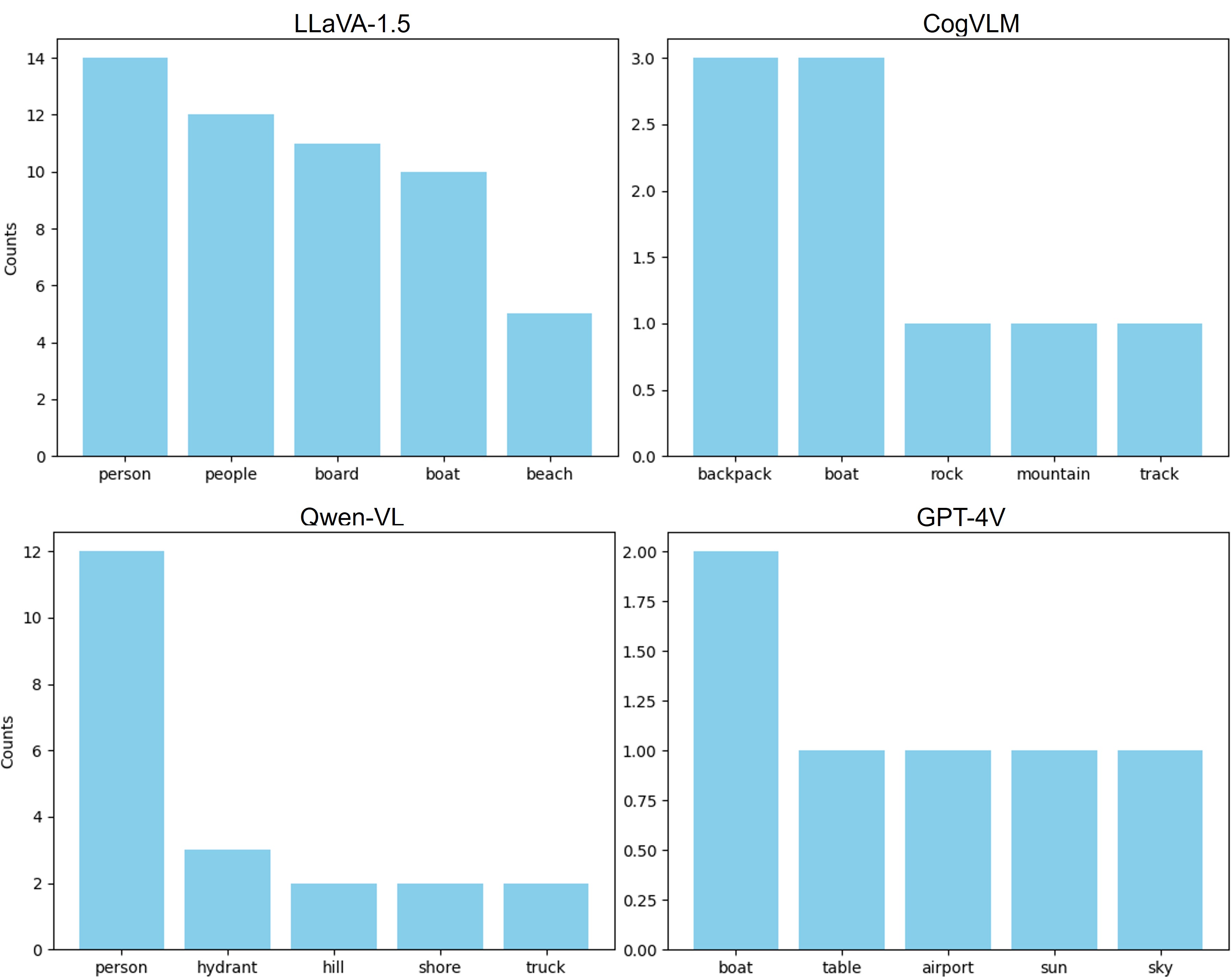}
    \caption{Statistics of the top 5 hallucinatory objects most frequently generated by various MLLMs when the object ``person'' is present.}
    \label{fig:asso}
\end{figure}

In addition, we conducted an analysis of the correlation between people and the objects that various MLLMs are prone to hallucinate. We selected data containing people or instances where MLLMs generated hallucinations of humans and then counted the objects most frequently hallucinated by MLLMs in this data. The results are shown in Figure~\ref{fig:asso}. It can be observed that objects such as boat, backpack, bench, etc., become the primary hallucinatory objects, and the occurrence of these objects is often related to humans. This once again confirms that the appearance of hallucinatory objects is influenced by the current scene in the image.

\begin{table*}[!ht]
    \centering
    \renewcommand{\arraystretch}{1.1}
    \setlength{\tabcolsep}{9pt}
    \scalebox{0.9}{
    \begin{tabular}{l | c c c c | c c c c | c}
    \hline
    \toprule
    \multicolumn{1}{l}{\multirow{2}{*}{\textbf{Model}}}&\multicolumn{4}{c}{\textsc{Generative Task}}&\multicolumn{4}{c}{\textsc{Discriminative Task}}&\multicolumn{1}{c}{\multirow{2}{*}{\textbf{\makecell{AMBER \\ Score}}}}\\
    \cmidrule(lr){2-5}
    \cmidrule(lr){6-9}
    \multicolumn{1}{c}{}&\multicolumn{1}{c}{CHAIR$_{\downarrow}$}&Cover$_{\uparrow}$&Hal$_{\downarrow}$&\multicolumn{1}{c}{Cog$_{\downarrow}$}&\multicolumn{1}{c}{Acc.}&P.&R.&\multicolumn{1}{c}{F1}&\multicolumn{1}{c}{}\\
    \midrule
    MiniGPT-4 7B&13.6&63.0&65.3&11.3&63.6&90.5&50.4&64.7&75.6\\
    MiniGPT-4 13B&14.3&63.6&60.3&9.6&64.9&90.1&52.9&66.7&76.2\\
    \midrule
    LLaVA-1.5 7B&7.8&51.0&36.4&4.2&72.0&93.2&62.4&74.7&83.5\\
    LLava-1.5 13B&7.0&51.8&33.1&3.3&72.2&96.0&59.0&73.1&83.1\\
    \midrule
    MiniGPT-v2 stage2&35.0&22.1&57.6&8.3&37.5&72.3&9.4&16.6&40.8\\
    MiniGPT-v2 stage3&32.8&31.6&67.4&10.4&36.2&74.2&5.8&10.8&39.0\\
    \bottomrule
    \hline
    \end{tabular}
    }
    \caption{Results of the ablation experiment for complete metrics.}
    \label{tab:llm_size_all}
\end{table*}

\begin{table}[!ht]
 \centering
  \renewcommand{\arraystretch}{1.1}
    \setlength{\tabcolsep}{6pt}
    \scalebox{0.92}{
  \begin{tabular}{l | c c c c} 
  \hline
    \toprule 
    \multicolumn{1}{l}{\multirow{2}{*}{\textbf{Model}}}&\multicolumn{2}{c}{Yes Ratio}&\multicolumn{2}{c}{No Ratio}\\
    \cmidrule(lr){2-3}
    \cmidrule(lr){4-5}
    \multicolumn{1}{c}{}&\multicolumn{1}{c}{All.}&w/ Hal.&All.&w/o Hal.\\
    \midrule
    mPLUG-Owl&92.4&43.3&7.6&0.9\\
    LLaVA&77.7&32.4&22.3&4.7\\
    MiniGPT-4&82.1&35.2&17.9&3.1\\
    CogVLM&71.8&27.5&28.2&5.7\\
    LLaVA-1.5&71.0&24.5&29.0&3.5\\
    mPLUG-Owl2&69.6&22.3&28.2&4.3\\
    InstructBLIP&49.5&11.7&50.5&12.2\\
    Qwen-VL&41.8&12.2&58.2&20.5\\
    GPT-4V&39.7&4.7&60.3&15.0\\
    \bottomrule
    \hline
  \end{tabular}}
  \caption{The ratio of MLLMs' response with ``Yes'' or ``No'' on discriminative task, where w/ Hal and w/o Hal represent the ground truth for queries with ``Yes'' and ``No'', respectively.}
  \label{tab:yesno_all}
\end{table}

\subsubsection{Ablation Results}

In Section~\ref{ablation_section}, we explored several factors that may influence hallucinations. Due to space constraints, we only provided results for some indicators. In this subsection, we will delve into a more comprehensive analysis of the complete results.

\begin{figure*}[!ht]
    \centering
    \includegraphics[width=0.85\textwidth]{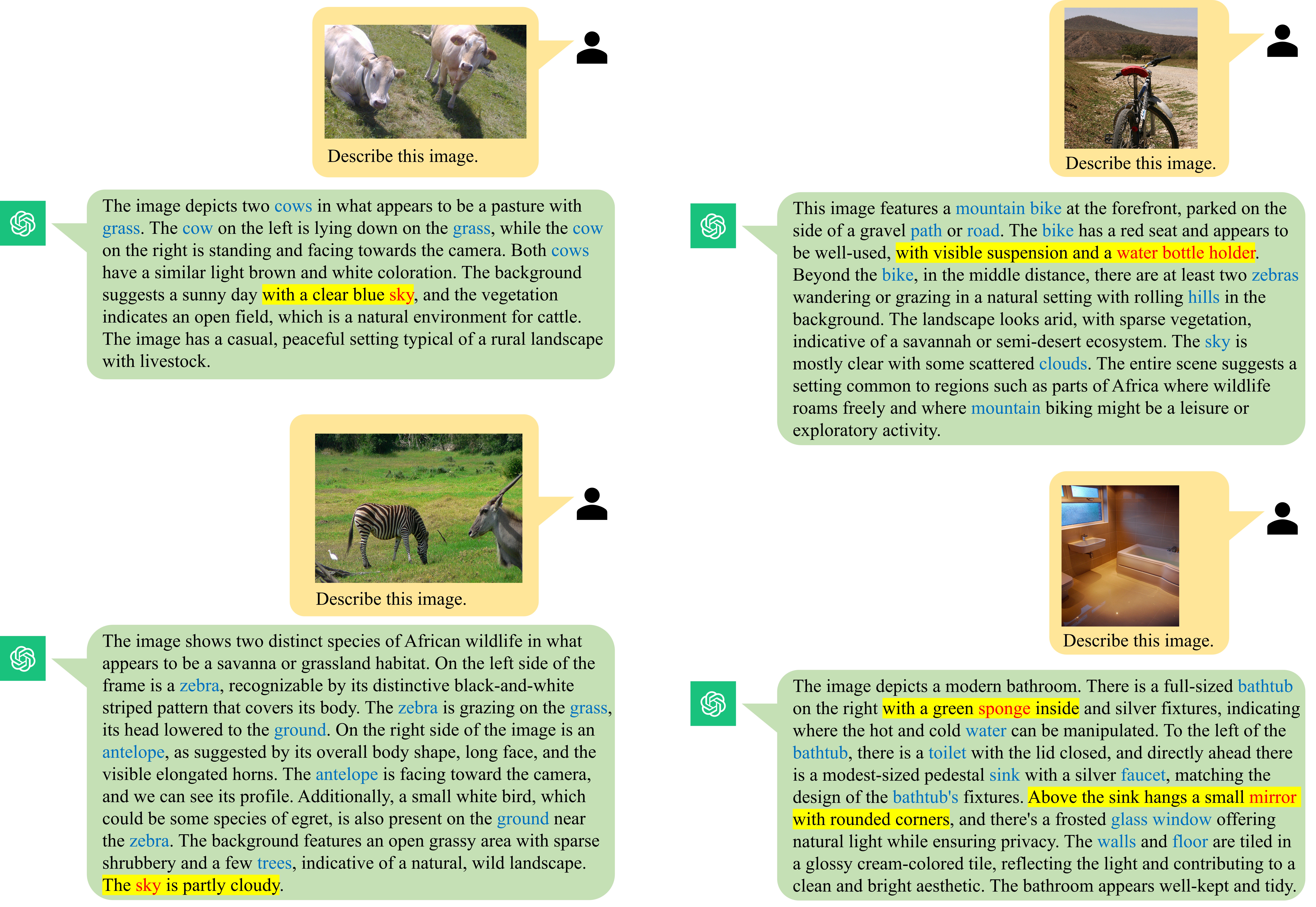}
    \caption{Some examples from AMBER that cause GPT-4V to generate hallucinations. The \hl{highlight} fonts represent the hallucinations. The \textcolor{red}{red} fonts and \textcolor{blue}{blue} fonts represent what AMBER detected w/ and w/o hallucinations, respectively.}
    \label{fig:gpt_case}
\end{figure*}

\noindent \textbf{The Scales of LLMs} \par
The results are shown in Table~\ref{tab:llm_size_all}. For MiniGPT-4, the increase in the parameters of LLM led to an increase in hallucinations on generative task. However, it also exhibited superior performance in terms of object coverage and the proportion of hallucination occurrences. Overall, there was a certain improvement in generative quality without a significant increase in hallucinations. on discriminative task, MiniGPT-4 achieved comprehensive improvements.

For LLaVA-1.5, increasing the parameters of LLM resulted in a comprehensive reduction of hallucinations on generative task, and at the same time, entity coverage also improved. It is evident that the size of LLM has a significant impact on generative task. However, on discriminative task, the 13B model performed poorly. This could be because discriminative task have lower requirements for the capabilities of LLM compared to generative task, so increasing the parameter count did not lead to an improvement in performance.\\

\noindent \textbf{Training Data} \par

In Table~\ref{tab:llm_size_all}, we present the results of the training data ablation experiment on MiniGPT-v2 for complete metrics. The trends observed in \textit{Hal} and \textit{Cog} align with the analysis presented in the main text.

\subsubsection{Cases of Hallucination in GPT-4V}

Figure~\ref{fig:gpt_case} shows some examples of GPT-4V generating hallucinations with the evaluation of AMBER. From the cases, it can be observed that even GPT-4V exhibits absurd illusions in AMBER. And we can see that AMBER successfully and accurately detects the hallucinations.

\end{document}